\documentclass[letterpaper, 10 pt, conference]{ieeeconf}  

\IEEEoverridecommandlockouts                              

\usepackage[font=footnotesize]{caption}
\usepackage{subcaption}
\usepackage{epsfig} 
\usepackage{times} 
\usepackage{amsmath} 
\usepackage{amssymb}  

\let\labelindent\relax
\usepackage{enumitem}
\usepackage{bm}
\let\mathcal\undefined
\DeclareMathAlphabet{\mathcal}{OMS}{cmsy}{m}{n}

\DeclareSymbolFont{Xlargesymbols}{OMX}{cmex}{m}{n}
\DeclareMathSymbol{\Xsum}{\mathop}{Xlargesymbols}{80}

\usepackage{tabularx, booktabs}
\usepackage{multirow}
\newcolumntype{Y}{>{\centering\arraybackslash}X}

\usepackage{placeins} 
\usepackage{flafter}  

\usepackage{lipsum}
\usepackage{color}
\usepackage[nospace]{cite}

\makeatletter
\let\NAT@parse\undefined
\makeatother

\usepackage{threeparttable}
\usepackage{url}
\usepackage[hidelinks]{hyperref}

\title{\LARGE \bf
Loosely-Coupled Semi-Direct Monocular SLAM
}

\author{Seong Hun Lee and Javier Civera
\thanks{Seong Hun Lee is with I3A, University of Zaragoza, 50018 Zaragoza, Spain
        (phone: +34 65 463 7956, e-mail: {\tt\small seonghunlee@unizar.es})
		        }
\thanks{ Javier Civera is with I3A, University of Zaragoza, 50018 Zaragoza, Spain
        (phone: +34 87 655 5554, e-mail: {\tt\small jcivera@unizar.es)}}
}

\begin{document}

\onecolumn

\begin{center}
\ %

This paper has been accepted for publication in \textit{IEEE Robotics and Automation Letters.}

\vspace{2em}

DOI: \href{http://doi.org/10.1109/LRA.2018.2889156}{10.1109/LRA.2018.2889156}

IEEE Xplore: \url{http://ieeexplore.ieee.org/document/8584894/}
\end{center}

\vspace{3em}

\textcopyright 2018 IEEE. Personal use of this material is permitted. Permission from IEEE must be obtained for all other uses, in any current or future media, including reprinting /republishing this material for advertising or promotional purposes, creating new collective works, for resale or redistribution to servers or lists, or reuse of any copyrighted component of this work in other works.
\twocolumn
\newpage

\maketitle
\thispagestyle{empty}
\pagestyle{empty}

\begin{abstract}
    We propose a novel semi-direct approach for monocular simultaneous localization and mapping (SLAM) that combines the complementary strengths of direct and feature-based methods.  
    The proposed pipeline loosely couples direct odometry and feature-based SLAM to perform three levels of parallel optimizations: (1) photometric bundle adjustment (BA) that jointly optimizes the local structure and motion, (2) geometric BA that refines keyframe poses and associated feature map points, and (3) pose graph optimization to achieve global map consistency in the presence of loop closures.
    This is achieved in real-time by limiting the feature-based operations to marginalized keyframes from the direct odometry module.
    Exhaustive evaluation on two benchmark datasets demonstrates that our system outperforms the state-of-the-art monocular odometry and SLAM systems in terms of overall accuracy and robustness.
\end{abstract}

\section{INTRODUCTION}
\label{sec:intro}
\begin{figure}[t]
 \centering
 \includegraphics[width=0.48\textwidth]{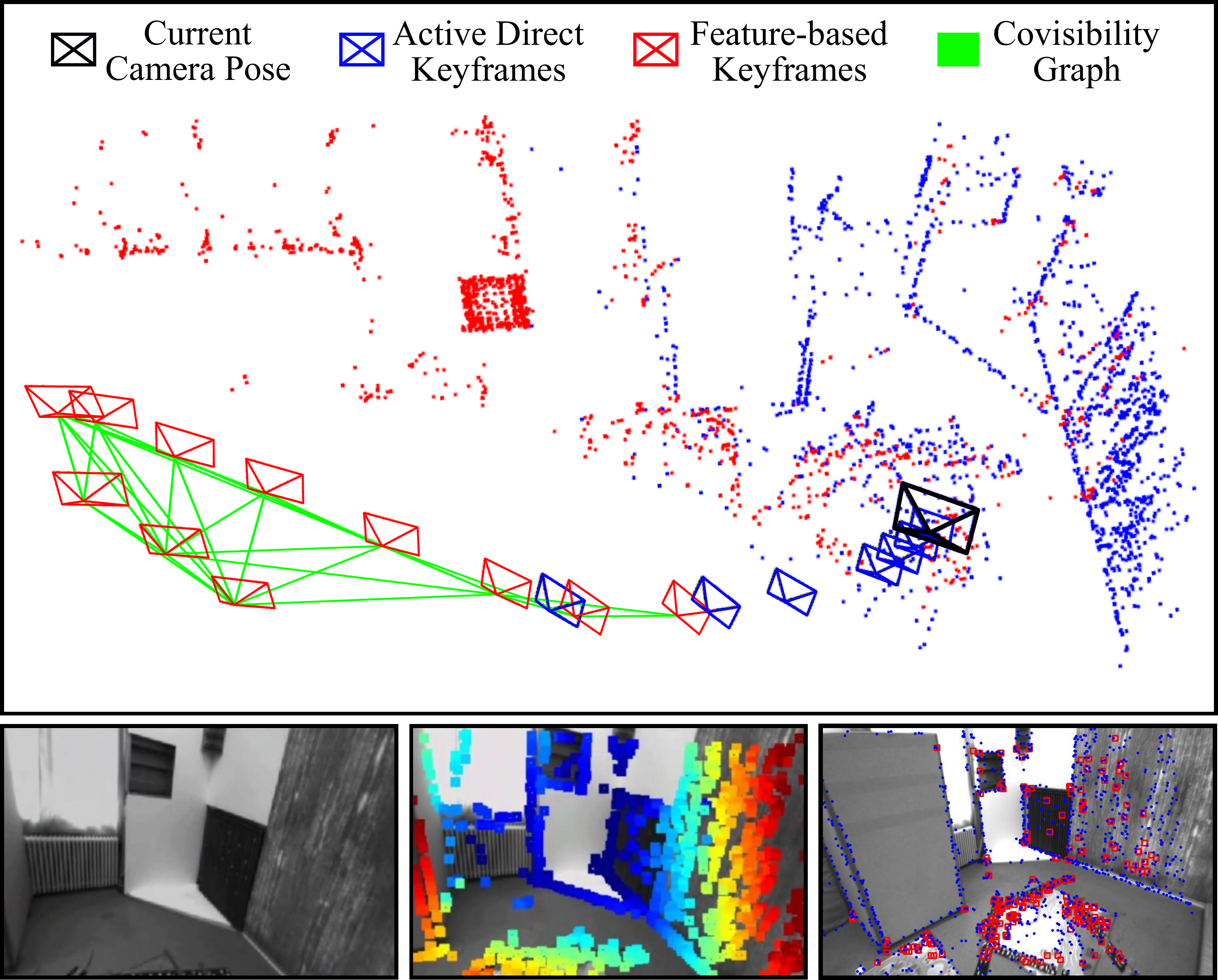}
\caption{
\textbf{Top:} We combine a direct and a feature-based method for monocular SLAM: the former is used for tracking and reconstructing a short-term local map (blue), and the latter for building a reusable global map (red and green).
\textbf{Bottom:} (from left to right) the current frame, the latest direct keyframe with color-coded depths, and the latest feature-based keyframe with the matched features (red) and the projection of direct map points (blue).
 }
\label{fig:map}
\vspace{-1em}
\end{figure}
Real-time visual odometry (VO) and simultaneous localization and mapping (SLAM) play an important role in many emerging technologies, such as autonomous ground/air vehicles \cite{autonomous_car, stability_scale} and virtual/augmented reality \cite{ptam}. 
In particular, monocular methods have drawn significant attention due to their minimal hardware constraints.

Traditional algorithms relied heavily on feature extraction and matching to estimate structure and motion \cite{mono-slam,ptam}.
In recent years, however, direct methods have gained rapidly increasing popularity \cite{lsd-slam, dso}. 
In contrast to feature-based ones, direct methods are capable of leveraging raw photometric information from a chosen set of pixels in the image.
This removes the need for costly per-frame feature extraction and matching.
Also, they are shown to be relatively more robust in low-texture scenes \cite{dso}.
 
Although direct methods have their own merits in several aspects, they inevitably miss certain benefits of salient features.
For example, feature descriptors such as SIFT \cite{sift} or ORB \cite{orb} have a high degree of invariance to viewpoint and illumination changes, and they can be matched over wide baselines. 
Such properties are favorable for tracking large inter-frame motions and recognizing revisited places.
Recent studies indeed confirm that direct and feature-based methods have their own strengths and weaknesses in respective areas \cite{dso,dso-vs-orb}. 
Semi-direct methods such as \cite{svo} attempt to take advantage of such complementary characteristics by incorporating ideas from both direct and feature-based methods.

In this paper, we propose a novel semi-direct approach for monocular SLAM that inherits both the robustness of direct VO and the map-reusing capability (e.g., loop closure) of feature-based SLAM. 
Our contribution is a \textit{loose coupling} between direct and feature-based algorithms such that:
\begin{enumerate}
\itemsep0em
    \item Locally, a direct method is used to track the camera pose rapidly and robustly with respect to a locally accurate, short-term, semi-dense map.
    \item Globally, a feature-based method is used to refine the keyframe poses, perform loop closures, and build a globally consistent, long-term, sparse feature map that can be reused. 
\end{enumerate}
This strategy allows us to complement the weaknesses of each method without compromising their real-time efficiency and performance.
We implement our approach on top of DSO \cite{dso} and ORB-SLAM \cite{orb-slam}, respectively the state-of-the-art in direct and feature-based methods, and demonstrate that our system outperforms both of them on two public benchmark datasets.
Fig. \ref{fig:map} shows an example snapshot of the estimated camera trajectory and associated scene reconstruction using our method.
The full reconstruction process is demonstrated in the accompanying video:
\vspace{-0.3em}
\begin{center}
 \url{https://youtu.be/j7WnU7ZpZ8c}
\end{center}
\vspace{-0.3em}
We make our implementation publicly available at:
\vspace{-0.3em}
\begin{center}
 \url{https://github.com/sunghoon031/LCSD_SLAM}
\end{center}


\begin{figure*}[t]
 \centering
 \vspace{5pt}
 \includegraphics[width=\textwidth]{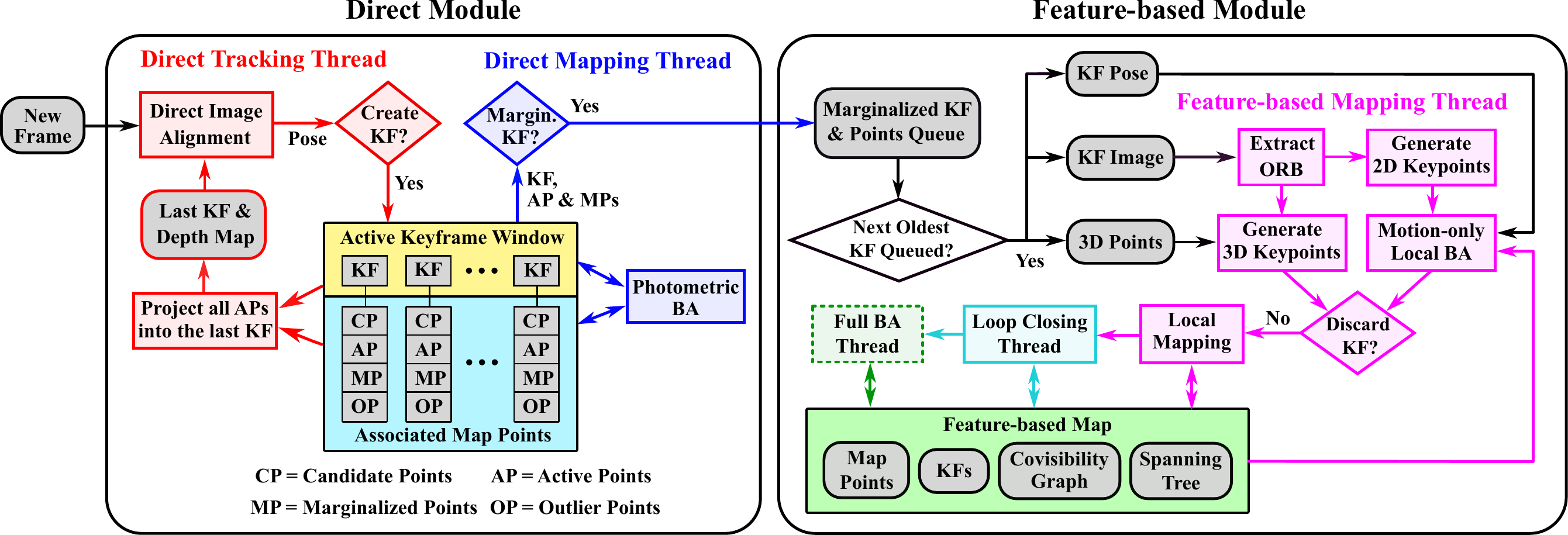}
\caption{Our pipeline consists of two modules operating in parallel: One is a direct module that tracks every new frame with respect to the last keyframe and performs windowed photometric BA. The other is a feature-based module that reconstructs a globally consistent map and keyframe trajectory using the marginalized information from the direct module.}
\label{fig:pipeline}
\vspace{-1em}
\end{figure*}

\vspace{-0.3em}
\section{RELATED WORK}
\label{sec:related}

Modern keyframe-based VO/SLAM systems can be categorized into three classes:

\textbf{(1) Feature-based:} Feature-based (or indirect) methods recover both camera pose and scene structure by matching features and performing \textit{geometric} bundle adjustment (BA) that minimizes the reprojection error.
PTAM \cite{ptam} is one of the most representative systems of this type, where it was first proposed to split tracking and mapping into two parallel threads. 
At present, ORB-SLAM \cite{orb-slam} is arguably the best-performing feature-based system.
Based on multiple successful ideas of PTAM and others \cite{bags_of_binary_words, scale-drift-aware, double_window}, ORB-SLAM uses ORB features \cite{orb} to perform tracking, mapping, relocalization and loop closing in a scalable manner. 
In \cite{orb-semidense}, an extension of ORB-SLAM was proposed to generate a semi-dense reconstruction of the scene.
This last method, however, does not use the resulting semi-dense map for tracking.

\textbf{(2) Direct:} Direct methods estimate structure and motion by minimizing the photometric error (i.e., intensity difference) between corresponding pixels in images \cite{about_direct_methods}.
Unlike most feature-based methods, direct methods are not limited to a sparse map and can maintain either a sparse \cite{dso}, semi-dense \cite{semidense,lsd-slam} or dense \cite{dtam,multilevel} map in real-time. 
Currently, the best-performing odometry system is DSO \cite{dso} which performs \textit{photometric} BA to jointly optimize camera intrinsics, extrinsics and inverse depths of sparse (or semi-dense) points in a sliding window fashion.
In \cite{sparse_vs_dense}, it was found that, for a small number of map points (e.g., $< 1000$), such joint optimization tends to be more accurate than alternating tracking and mapping as in, for example, LSD-SLAM \cite{lsd-slam}.
DSO is also the first work to demonstrate the benefits of photometric calibration \cite{photo_calibration} in direct methods.
However, it is subject to drift over time as it is a pure odometry method and does not reuse the map points that leave the field of view (FOV).
Gao et al. \cite{ldso} recently presented a modified DSO with the loop-closing capability similar to LSD-SLAM.
Unlike our method, however, their system does not produce a reusable sparse feature map which is useful for applications such as global relocalization, fixed-map tracking and collaborative mapping.

\textbf{(3) Semi-Direct:} Semi-direct (or hybrid) methods estimate camera poses using both direct and feature-based methods. 
For example, SVO \cite{svo} performs direct sparse image alignment to estimate the initial guess of the camera pose and feature correspondences. 
Afterwards, it performs geometric BA to refine the pose and structure. 
It was shown in \cite{remode} that SVO could also be used for dense mapping.
In \cite{svo2}, several improvements to the original SVO were proposed. 
Although this system is highly efficient, it was shown to be less robust \cite{svo2,dso-vs-orb} than ORB-SLAM \cite{orb-slam} and DSO \cite{dso}.
In \cite{li_semidirect} and \cite{semidirect_rgbd}, similar approaches to SVO were proposed for monocular visual(-inertial) and RGB-D SLAM, respectively.
Both methods adopt a direct method for tracking and a feature-based method for keyframe pose refinement, mapping and loop closing.
At the end of Section \ref{sec:overview}, we briefly discuss how our method differs from these existing semi-direct methods.
In \cite{krombach} and \cite{kim_hybrid}, different semi-direct approaches were proposed for stereo odometry.
Both methods use feature-based tracking to obtain a motion prior, and then perform direct semi-dense or sparse alignment to refine the camera pose.
While they were respectively shown to perform well against large inter-frame motions and illumination changes, they do not utilize the robustness of direct tracking in low-texture scenes. 
\vspace{-0.3em}
\section{SYSTEM OVERVIEW}
\label{sec:overview}
Fig. \ref{fig:pipeline} illustrates our proposed semi-direct pipeline. 
The idea is to combine the currently best-performing feature-based and direct algorithms, namely ORB-SLAM \cite{orb-slam} and DSO \cite{dso}, with some modifications.
To achieve real-time performance, we take inspiration from SVO \cite{svo} and apply a direct method to quickly track each frame and provide an initial seed for feature-based map optimization. 
Specifically, we use DSO to achieve real-time tracking and a modified version of ORB-SLAM to build a globally consistent map at a slower rate with marginalized keyframes from DSO. 
This is shown in Fig. \ref{fig:pipeline} as a direct and a feature-based module, respectively.
As the two separate asynchronous modules exchange information without sharing states, this approach is considered \textit{loosely coupled}.

Our system architecture involves three different layers of optimization windows. 
At the most local level, a sliding window of keyframes and map points are photometrically bundle adjusted to obtain an accurate representation of the surrounding environment.
New frames are tracked using direct image alignment \cite{lucas-kanade-unifying1} with respect to the last keyframe and its depth map created by projecting active points in the window (see Fig. \ref{fig:map}). 

When a keyframe is marginalized from the direct module, its image and pose information is sent to the feature-based module, along with the map points within its FOV.
The feature-based module extracts ORB descriptors from these keyframes and refines their poses with respect to the local feature map using motion-only BA.
Some of these keyframes and map points are added to the local map and geometrically bundle adjusted for optimal accuracy.
This process of feature-based mapping corresponds to the mid-level optimization.

Finally, at the most global level, a pose graph optimization \cite{scale-drift-aware} is performed over Sim(3) constraints after each loop detection.
Afterwards, a full BA \cite{orb-slam2} optimizes all keyframes and feature points in the map for global consistency.

The key difference between our approach and SVO \cite{svo} (or the semi-direct methods in \cite{li_semidirect,semidirect_rgbd}) is that we maintain in parallel two separate maps, each in the direct and the feature-based module.
This allows us to utilize a locally accurate semi-dense map for fast and robust tracking, as well as a globally consistent sparse feature map for long-term reuse (e.g., loop detection and closure).
\vspace{-0.3em}
\section{NOTATION}
\label{sec:notation}
Throughout the paper, we use bold lower- and upper-case letters for vectors ($\mathbf{v}$) and matrices ($\mathbf{M}$), and light lower- and upper-case letters for scalars ($s$) and scalar functions ($F$), respectively.
The intensity image is denoted by $I:\Omega \mapsto \mathbb{R}$ where $\Omega \subset \mathbb{R}^2$ is the image domain. 
We denote the camera intrinsic parameters with $\mathbf{c}$, and corresponding camera projection and back-projection functions with $\bm{\Pi}_\mathbf{c}:\mathbb{R}^3\mapsto \Omega$ and $\bm{\Pi}^{-1}_\mathbf{c}:\Omega \times \mathbb{R}\mapsto \mathbb{R}^3$, respectively.
Camera poses are represented as either rigid body or similarity transformation matrices $\mathbf{T}_{iw} \in \text{SE(3)}$ or $\mathbf{S}_{iw} \in \text{Sim(3)}$ that transform a point from the world frame to frame $i$.
We use $\mathcal{P}_i$ to denote the set of map points belonging to keyframe $i$ and $\text{obs}(\mathbf{p})$ to denote the set of keyframes in which the point $\mathbf{p}$ is visible.
The Euclidean and Huber norms \cite{huber} are denoted by $\left\Vert \cdot \right\Vert_2$ and $\left\Vert \cdot \right\Vert_\gamma$ respectively.
The operator $\boxplus$ is defined as a simple addition for Euclidean parameters and a left multiplication for the pose, i.e., for Lie-algebra $\mathfrak{se}$(3) elements in twist coordinates $\mathbf{x} \in \mathbb{R}^6$, $\mathbf{x} \boxplus \mathbf{T}:= \text{exp}_\text{SE(3)}\left(\mathbf{x}\right)\mathbf{T}$.
We use the same notation as in \cite{scale-drift-aware} for the exponential and logarithmic mapping for SE(3) and Sim(3).
\vspace{-0.3em}
\section{DIRECT MODULE}
\label{sec:direct}
We use the original implementation of DSO \cite{dso} as our direct module, which is responsible for initial map bootstrapping, real-time camera tracking and local mapping.
In this section, we describe the windowed optimization and marginalization scheme of DSO.
The reader is referred to the original work \cite{dso} for details on direct tracking and other front-end operations.
\vspace{-0.3em}
\subsection{Windowed Photometric Bundle Adjustment}
\label{subsec:windowed_photo_BA}
When a point $\mathbf{p}$ in a reference frame $I_i$ is observed in another frame $I_j$, the photometric error is defined as the weighted SSD over the 8-point neighborhood pixels $\mathcal{N}_{\mathbf{p}}$ as proposed in \cite{dso}: 
\begin{equation}
\label{eq:photo_residual}
    E^{\mathbf{p}}_{ij} := \Xsum_{\Tilde{\mathbf{p}} \in \mathcal{N}_\mathbf{p}} \omega_{\Tilde{\mathbf{p}}} \left\Vert I_j\left[\Tilde{\mathbf{p}}'\right] -b_j-\frac{t_je^{a_j}}{t_ie^{a_i}}\left(I_i\left[\Tilde{\mathbf{p}}\right] -b_i\right)\right\Vert_\gamma
\end{equation}
\vspace{-0.5em}
\begin{equation}
    \text{with} \quad \omega_{\Tilde{\mathbf{p}}} := \frac{c^2}{c^2+\left\Vert \nabla I_i(\Tilde{\mathbf{p}})\right\Vert^2_2},
\end{equation}
\vspace{-0.5em}
\begin{equation}
    \Tilde{\mathbf{p}}'=\mathbf{\Pi}_{\mathbf{c}}\left(\mathbf{T}_{jw}^{-1}\mathbf{T}_{iw}\mathbf{\Pi}^{-1}_{\mathbf{c}}\left(\Tilde{\mathbf{p}},d_{\mathbf{p}}\right)\right)
\end{equation}
where $t$, $a$ and $b$ are exposure time and affine brightness function parameters, and $d_{\mathbf{p}}$ is the inverse depth \cite{inverse_depth} of $\mathbf{p}$ in the reference frame $I_i$.
The weight $w_{\Tilde{\mathbf{p}}}$ down-weights high-gradient pixels with some constant $c$.
The total energy function to be minimized is given by the full photometric error plus a prior pulling the affine brightness parameters to zero:
\vspace{-0.3em}
\begin{equation}
\label{eq:total_energy}
   E_\text{photo} := \Xsum_{i \in \mathcal{F}} \Xsum_{\mathbf{p}\in \mathcal{P}_i}\Xsum_{j\in \text{obs}(\mathbf{p})} E^{\mathbf{p}}_{ij}+\Xsum_{i\in \mathcal{F}}\left(\lambda_aa^2_i+\lambda_bb_i^2 \right),
   \vspace{-0.3em}
\end{equation}
where $\mathcal{F}$ denotes the set of all frames in the window.
When exposure times are known, we set $\lambda_a$ and $\lambda_b$ to some constant values.
Otherwise, we set $\lambda_a=\lambda_b=0$ and $t_i=t_j=1$ in \eqref{eq:photo_residual}.
The optimization is performed using an iteratively reweighted Gauss-Newton or Levenberg-Marquardt algorithm in a coarse-to-fine scheme.
The update equation is given by
\vspace{-0.5em}
\begin{equation}
    \delta \bm{\xi} =-\left(\mathbf{J}^T\mathbf{W}\mathbf{J}\right)^{-1}\mathbf{J}^T\mathbf{W}\mathbf{r} \quad \text{and} \quad \bm{\xi}^{\text{new}} \leftarrow \delta \bm{\xi} \boxplus \bm{\xi},
    \vspace{-0.3em}
\end{equation}
where $\mathbf{r}$ is the stacked vector of residuals, $\mathbf{J}$ is its Jacobian and $\mathbf{W}$ is the diagonal weight matrix.
The state variable $\bm{\xi}$ contains all the active variables in the window, i.e., camera poses, affine brightness parameters, inverse depths and camera intrinsics.
\vspace{-0.3em}
\subsection{Marginalization}
\label{subsec:marginalization}
The size of the optimization window is kept bounded by marginalizing the least useful keyframes and points using the Schur complement \cite{dso,keyframe-based-leutenegger}.
Points are marginalized if they are not observed in the latest two keyframes or their host keyframe is marginalized.
Keyframes (other than the latest two) are marginalized if either less than 5\% of its points are visible in the latest keyframe, or if it has the highest ``distance score" when the window contains more than a certain number of keyframes.
We refer to the original work \cite{dso} for the computation of this heuristic score.
\vspace{-0.2em}
\section{FEATURE-BASED MODULE}
\label{sec:feature-based}
When a keyframe is marginalized from the direct module, the feature-based module receives its image and pose information, as well as the 3D locations of both active and marginalized map points within its FOV.
This information is then used for feature-based pose refinement, mapping and loop closing.
Note that the marginalization strategy in Section \ref{subsec:marginalization} does not necessarily marginalize the oldest keyframe in the window.
To avoid temporal inconsistency in such cases, we store the marginalized keyframes and points in a queue and wait until the next oldest keyframe is marginalized.
If more than five keyframes are queued and the next oldest keyframe is still active, we take its latest pose and points to proceed further instead of waiting. 
All optimizations are performed using the original implementation in ORB-SLAM \cite{orb-slam}, which is based on the Levenberg-Marquardt algorithm in g2o \cite{g2o}.
\vspace{-0.3em}
\subsection{Relative Scale and Initial Pose Estimation}
\label{subsec:initial_pose}
In our loosely-coupled approach, the direct and the feature-based modules maintain two separate maps.
Due to the scale ambiguity of a monocular system, the scales of these two maps drift over time and do not converge to the same value. 
Therefore, we continuously compute the relative scale using Sim(3) alignment \cite{arun,horn} between the 30 most recent keyframes in the feature-based module and their counterparts in the direct module.

Once the relative scale $s$ is known, the incremental transformation in the direct module can be scaled appropriately and used as an initial pose guess in the feature-based module:
Let $i$ and $j$ denote the previous and current keyframe.
Then,
\begin{equation}
\mathbf{T}_{jw|\text{F}}
=
\begin{bmatrix}
\mathbf{R}_{ji|\text{F}} & \mathbf{t}_{ji|\text{F}}\\
\mathbf{0}_{1\times3} & 1
\end{bmatrix}
\mathbf{T}_{iw|\text{F}} \ 
\end{equation}
\vspace{-1em}
\begin{align}
    \text{with} \ \ \mathbf{R}_{ji|\text{F}} &= \mathbf{R}_{ji|\text{D}}=\mathbf{R}_{jw|\text{D}}(\mathbf{R}_{iw|\text{D}})^\text{T} , \label{eq:derived1}\\
    \mathbf{t}_{ji|\text{F}} &= s\hspace{1pt}\mathbf{t}_{ji|\text{D}} = s\left(-\mathbf{R}_{ji|\text{D}}\mathbf{t}_{iw|\text{D}}+\mathbf{t}_{jw|\text{D}}\right), \label{eq:derived2}
\end{align}
where the subscripts $\text{D}$ and $\text{F}$ indicate the direct and the feature-based module, respectively.
For the derivation of \eqref{eq:derived1} and \eqref{eq:derived2}, please refer to the supplementary material available at our public source-code repository (see Section \ref{sec:intro}).

\begin{figure}[t]
 \centering
 \vspace{5pt}
 \includegraphics[width=0.48\textwidth]{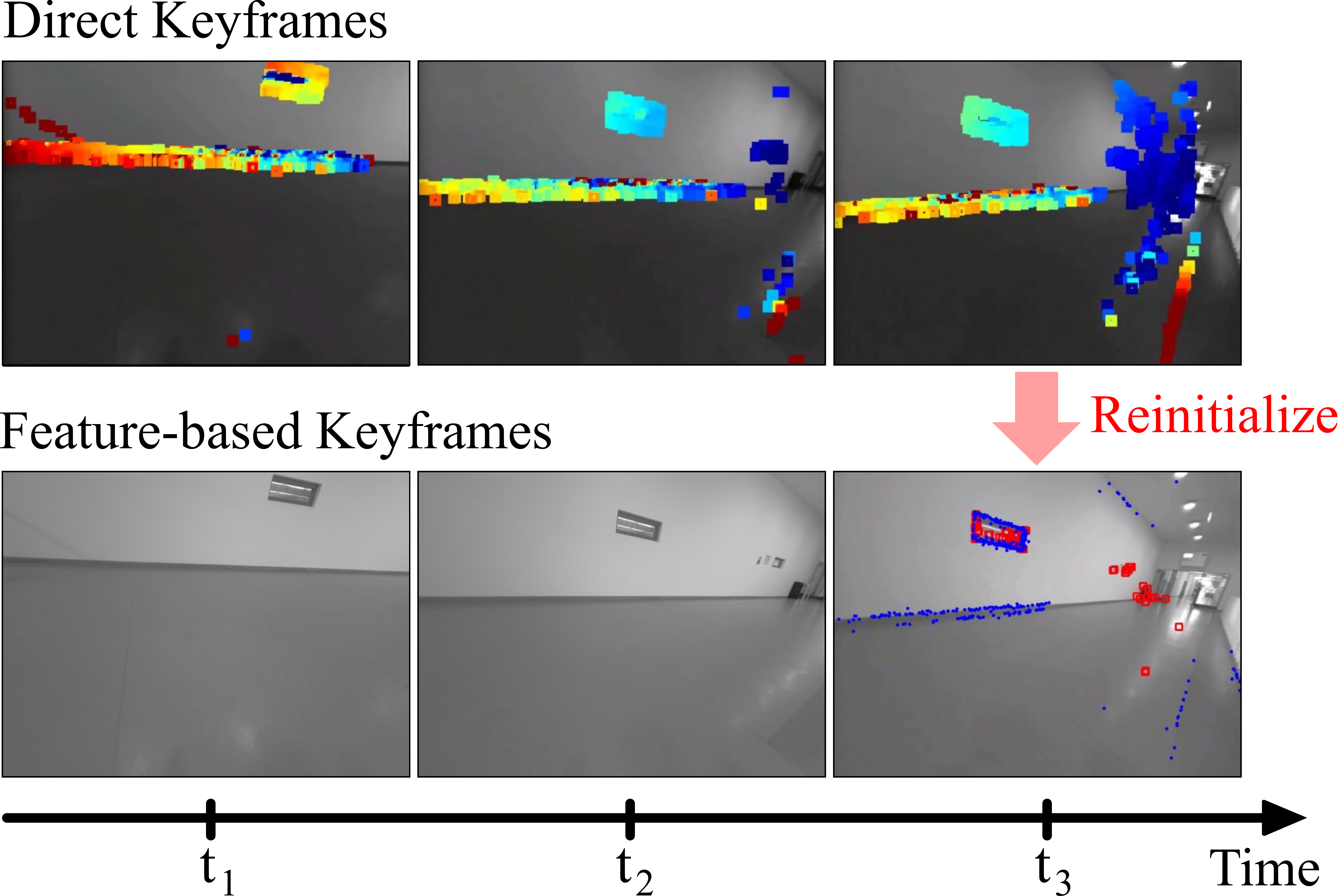}
\caption{\textbf{[TUM monoVO]} Failure recovery in \textit{sequence 40}: At $t_1$ and $t_2$, the feature-based module fails due to the lack of features in the scene, whereas the direct module is able to track the high-gradient pixels. At $t_3$, the scene now contains a sufficient number of features, and their depths can be initialized with the help of the direct module.}
\label{fig:reinit}
\vspace{-5pt}
\end{figure}

\vspace{-0.3em}
\subsection{3D Keypoints Generation}
\label{subsec:fb_init}
The map points from the direct module are used in two ways: (1) for creating an initial set of 3D keypoints to bootstrap the feature-based module, or (2) for adding more local map points to improve the tracking robustness.
Given the 2D position $\mathbf{p}$ of an ORB feature in frame $i$, we generate a 3D keypoint $\mathbf{x}_w$ in the world frame as
\begin{equation}
    \mathbf{x}_w=\mathbf{T}_{iw}^{-1}\bm{\Pi}^{-1}_{\mathbf{c}}\left(\mathbf{p},\frac{d_{\mathbf{p}}}{s}\right) \quad \text{with} \quad  d_{\mathbf{p}} = \frac{\Xsum_{k\in \mathcal{P}_{\mathbf{p}}} d_k /\sigma_k^2}{\Xsum_k 1/\sigma_k^2},
\end{equation}
where $s$ is the relative scale between the direct and the feature-based module (Section \ref{subsec:initial_pose}), and $\mathcal{P}_{\mathbf{p}}$ and $d_{k\in \mathcal{P}_{\mathbf{p}}}$ respectively denote the set of all map points whose projection in frame $i$ is equal to $\mathbf{p}$ and their inverse depths in frame $i$. 
We dilate the projection by two pixels to ensure a sufficient number of valid depths for the keypoints.
Note that the inverse depth $d_{\mathbf{p}}$ is computed as the inverse-variance weighted average, which is equivalent to the Kalman filter update with multiple measurements \cite{semidense}.

We found that extracting more features during slower camera motions often increases the number of covisibility links \cite{orb-slam} between the keyframes and improves mapping accuracy.
Therefore, we vary the number of features to extract per image as follows:
Let $f_\text{kf}$ be the keyframe addition frequency in the direct module.
Using this as the indicator of the relative camera speed, we set $N_\text{features}=2500$ if $f_\text{kf}< 4 \text{Hz}$ and $N_\text{features}=1500$ if $f_\text{kf}> 7 \text{Hz}$. 
Otherwise, we interpolate between the two bounds based on $f_\text{kf}$.

\vspace{-0.3em}
\subsection{Keyframe Pose Refinement and Failure Recovery}
\label{subsec:fb_refinement}
Once the direct module provides an initial pose estimate of a new keyframe (Section \ref{subsec:initial_pose}), we refine it using motion-only geometric BA with respect to the local feature map.
The total energy function is composed of the variance-normalized reprojection errors of the local map points:
\begin{gather}
\label{eq:reproj_error}
    E_\text{reproj}= \Xsum_{i\in \mathcal{F}_\text{local}} \Xsum_{\mathbf{x} \in \mathcal{P}_i} \Xsum_{j \in \text{obs}(\mathbf{x})} \left\Vert \frac{\mathbf{p}_{j, \mathbf{x}}-\bm{\Pi}_\mathbf{c}\left( \mathbf{T}_{iw} \mathbf{x}_w\right)}{\sigma^2_\mathbf{x}}\right\Vert_\gamma\\
    \text{with} \quad \sigma^2_{\mathbf{x}}:=\left(\lambda_\text{pyr}\right)^{2L_{\text{pyr}, \mathbf{x}}},
\end{gather}
where $\mathcal{F}_\text{local}$ denotes the set of all local keyframes, i.e., all keyframes sharing map points with the new keyframe and their neighbors in the covisibility graph \cite{orb-slam}, $\mathbf{p}_{j,\mathbf{x}} \in \mathbb{R}^2$ the match to the keypoint $\mathbf{x}$ in frame $j$, and $\sigma^2_{\mathbf{x}}$ the variance of the feature location in frame $i$.
This variance depends on the constant scale factor of the image pyramid  $\lambda_\text{pyr}$ ($>1$) and the pyramid level $L_{\text{pyr}, \mathbf{x}}$ at which the keypoint was detected.


When the feature-based module fails due to insufficient matches, we reinitialize the module, following the procedure explained in Section \ref{subsec:fb_init}.
This is illustrated in Fig. \ref{fig:reinit}.
Since our direct module is robust to low-texture scenes, we can rely on its tracking while the feature-based module is lost.
Then, as soon as we detect more features, we reinitialize the local map points using the depths from the direct module. 

\vspace{-0.3em}
\subsection{Feature-based Local Mapping and Loop Closing}
\label{subsec:fb_mapping_loop}
After generating 3D keypoints and refining the keyframe pose, we insert them in the feature-based map if the number of matches falls below 150 or more than three keyframes passed from the last insertion.
Once the keyframe and the points are added to the map, they are processed by the local mapping, as outlined in \cite{orb-slam}.
This includes the local geometric BA that minimizes \eqref{eq:reproj_error} to jointly optimize the current keyframe, its neighbors in the covisibility graph, and all the map points belonging to those keyframes.

In the loop closing thread, the place recognition module \cite{fast-relocal} based on DBoW2 \cite{bags_of_binary_words} detects large loops by querying the keyframe database.
Once a loop is detected, the keyframes and map points on each side of the loop are aligned and fused.
To correct the scale drift, pose graph optimization \cite{scale-drift-aware} is performed over the essential graph \cite{orb-slam}, minimizing
\begin{equation}
    E_\text{graph} = \Xsum_{(i,j) \in \mathcal{E}_\text{edge}}\left\Vert \text{log}_\text{Sim(3)}\left(\mathbf{S}_{ij,0} \ \mathbf{S}_{jw} \ \mathbf{S}^{-1}_{iw}\right)\right\Vert^2_2 ,
\end{equation}
where $\mathcal{E}_\text{edge}$ denotes the set of edges in the essential graph, and $\mathbf{S}_{ij,0}=\mathbf{S}_{iw,0}\mathbf{S}_{jw,0}^{-1}$ is the fixed similarity transformation (with the scale 1) between the frame $i$ and $j$ just prior to the pose graph optimization.
If the edge is created from a loop closure, this transformation is instead computed using the method of Horn \cite{horn}.
Finally, a full BA \cite{orb-slam2} is performed afterwards.

\vspace{-0.3em}
\subsection{Does Feature-based Mapping Always Improve Accuracy?}
\label{subsec:does_it_always_improve}
The answer is no. 
In general, the feature-based mapping described in the previous sections improves the accuracy if there is a loop closure or when the camera motion is mostly loopy, which enhances the covisibility of features in multiple keyframes and the reuse of map points.
However, we found that it actually causes more drift when the camera motion is mostly exploratory without loop closures (a similar finding was also reported in \cite{dso}). 

\begin{figure}[t]
 \centering
 \vspace{5pt}
 \includegraphics[width=0.35\textwidth]{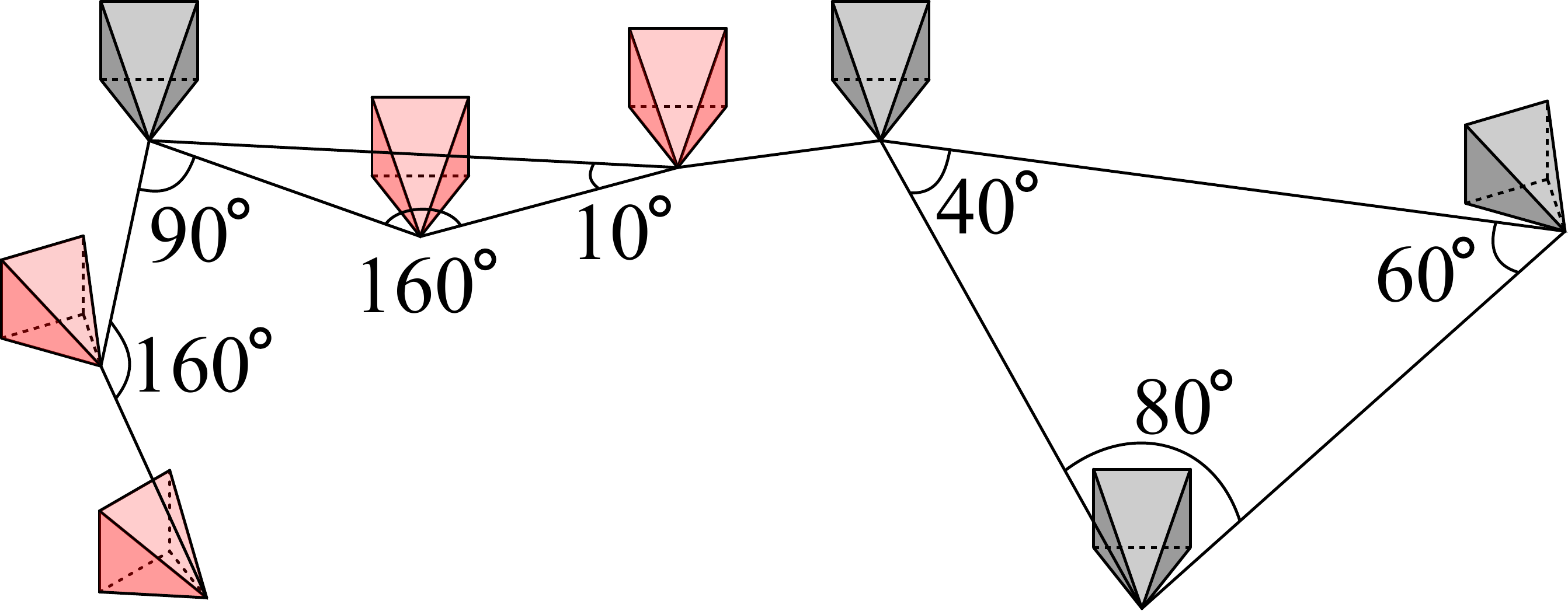}
\caption{Illustration of the keyframes with collinear covisibility links (red). None of their covisibility links form an angle between $30^\circ$ and $150^\circ$.}
\label{fig:collinear_covlinks}
\vspace{-5pt}
\end{figure}

We solve this by keeping two versions of the keyframe trajectory: one that is initially given by the direct module and the other modified by the feature-based module. 
We assume that the latter is more accurate if there is a loop closure or less than a quarter of all past keyframes have \textit{collinear covisibility links} at a given point in time. 
The covisibility links (i.e., 3D lines connecting the keyframe to its neighbors in the covisibility graph) are considered collinear if none of them form an angle between $30^\circ $ and $150^\circ $.
This is illustrated in Fig. \ref{fig:collinear_covlinks}.
We found that this method is especially effective for detecting exploratory translational motions.

While this strategy allows us to mitigate the odometry drift in exploratory situations, a more elegant solution would be to deal with the source of inaccuracy in the feature-based mapping that causes drift.
Yang et al. \cite{dso-vs-orb} suggests a few hints on how this could be done (e.g., careful point management and sub-pixel matching refinement), but it is still an open problem and remains for future work.

\section{EVALUATION}
\label{sec:evaluation}
\subsection{Evaluated Settings, Datasets and Methodology}
We implement our method using ROS\footnote{Robot Operating System, \url{http://www.ros.org/}.} and compare it against ORB-SLAM \cite{orb-slam} and DSO \cite{dso}.
We evaluate each algorithm in two different settings:
\begin{itemize}[leftmargin=*]
\itemsep0em
    \item \textbf{ORB-VO} and \textbf{ORB-SLAM}: 
    For the VO setting, we disable the loop closing thread.
    Relocalization is disabled for both settings to evaluate the tracking robustness.
    We do not apply photometric calibration \cite{photo_calibration}, as we found that it worsens feature extraction and matching (similar findings were also reported in \cite{dso,dso-vs-orb}).
    \item \textbf{DSO-default} and \textbf{DSO-reduced}: 
    We use the default and reduced settings provided in the original DSO implementation. 
    The only difference is that, for the reduced setting, we always resize the input images by half.
    Photometric calibration \cite{photo_calibration} is applied when available.
    \item \textbf{Ours-VO} and \textbf{Ours-SLAM}:
    The VO setting uses DSO-reduced and ORB-VO settings, whereas the SLAM setting uses DSO-reduced and ORB-SLAM settings.
    This means that the direct module processes photometrically calibrated images (if available) at half resolution, while the feature-based module processes photometrically non-calibrated images at full resolution.
    We found that using half resolution in the feature-based module significantly worsens the performance, which is in line with the observation in  \cite{photo_calibration}.
    For efficiency, we reduce the number of iterations in the local geometric BA by half.
\end{itemize}
We use two public benchmark datasets for evaluation:
\begin{enumerate}[leftmargin=*]
\itemsep0em
\item \textbf{EuRoC MAV dataset} \cite{euroc_mav_dataset}, which contains 11 indoor stereo sequences with $752\times480$ pixel resolution at 20 fps.
As in \cite{dso}, we crop the beginning and end of each sequence to disregard large occlusions due to the ground and aggressive motions meant for IMU initialization.
We evaluate the tracking accuracy using the absolute trajectory RMSE ($e_\text{ate}$) of keyframe poses after Sim(3) alignment with the ground truth.
Photometric calibration and exposure times are not available for this dataset.
\item \textbf{TUM monoVO dataset} \cite{photo_calibration}, which contains 50 in- and outdoor monocular sequences recorded at 20--50 fps.
We use $640\times480$ pixel resolution for undistorted images.
Since the full ground-truth data is not available for this dataset, the tracking accuracy is evaluated in terms of the alignment error ($e_\text{align}$) proposed in \cite{photo_calibration}.
The dataset also provides photometric calibration and exposure times.
\end{enumerate}
\begin{figure}[t]
 \centering
 \vspace{5pt}
 \includegraphics[width=0.44\textwidth]{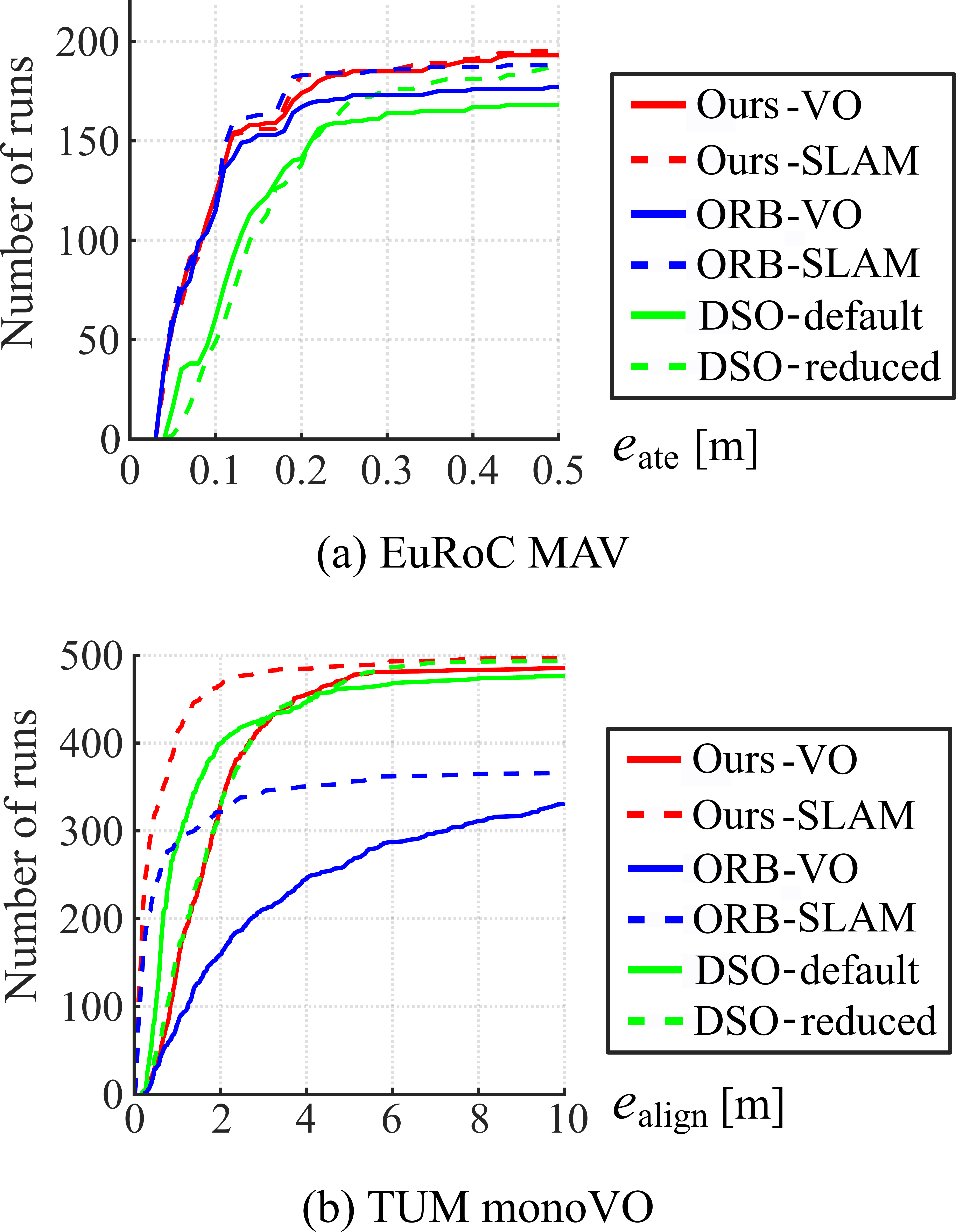}
\caption{Cumulative error plot aggregating \textbf{(a)} the absolute trajectory errors $e_\text{ate}$ [m] over all runs of all EuRoC MAV sequences and \textbf{(b)} alignment errors $e_\text{align}$ [m] over all runs of all TUM monoVO sequences. 
The closer the curve is to the y-axis, the higher the accuracy, as it means more runs with low errors. 
The farther the end of the curve is from the x-axis, the higher the robustness, as it means more runs without tracking failures.}
\label{fig:cumulative}
\vspace{-5pt}
\end{figure}
\begin{figure}[t]
 \centering
 \vspace{5pt}
 \includegraphics[width=0.48\textwidth]{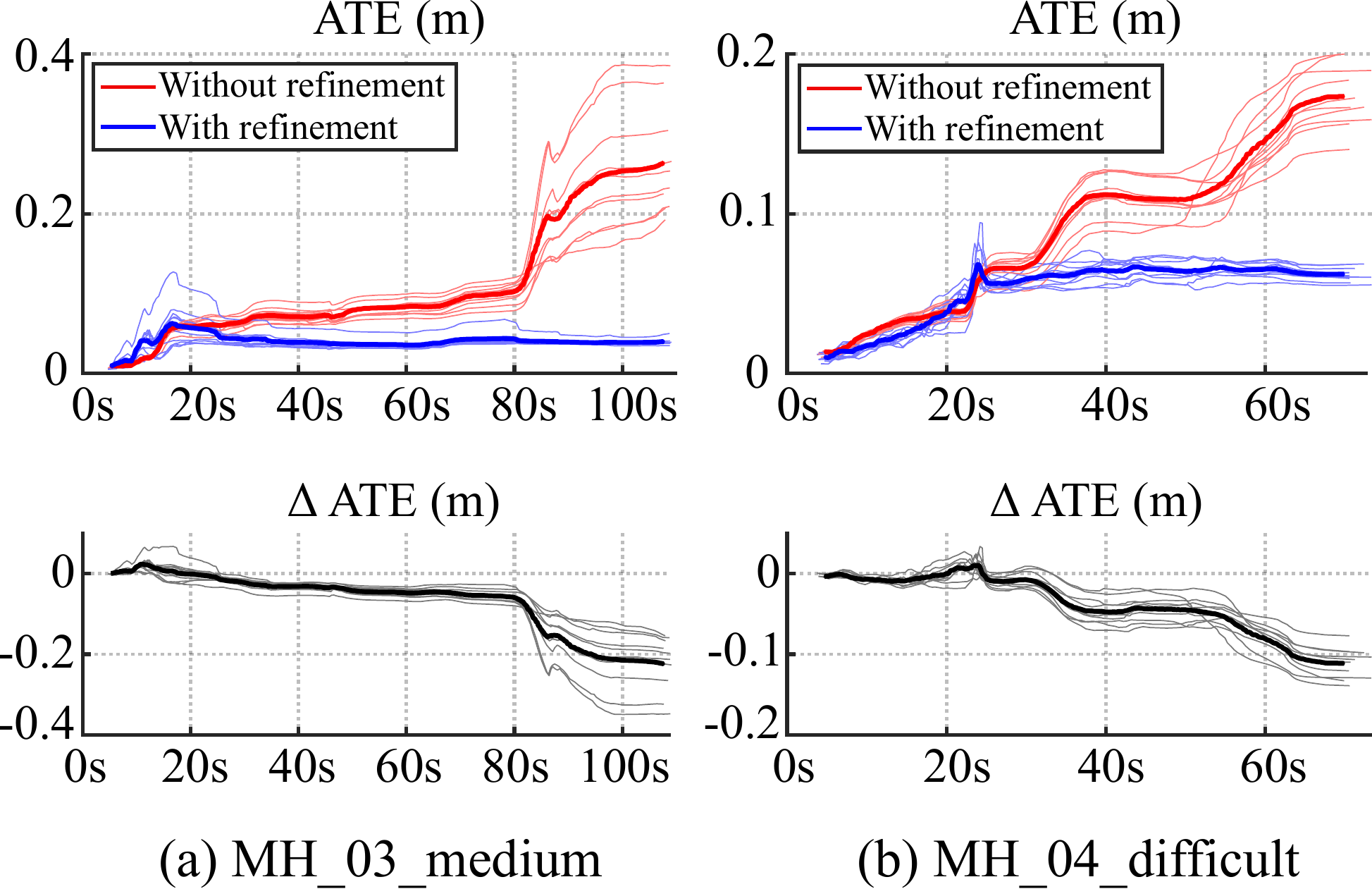}
\caption{\textbf{[EuRoC MAV] Top:} Time evolution of the ATEs with and without the feature-based refinement (Section \ref{subsec:fb_refinement} and \ref{subsec:fb_mapping_loop}). 
The loop-closure detection is disabled.
The average of 10 independent runs is shown in bold.
\textbf{Bottom:}  The ATE difference over time, i.e., the gap between the red and the blue curves.
}
\label{fig:evolution}
\vspace{-5pt}
\end{figure}
To account for non-deterministic behaviour, we run each method 10 times.
This amounts to 220 runs for the EuRoC MAV dataset (i.e., 110 runs for each left and right camera) and 500 runs for the TUM monoVO dataset.
On the EuRoC MAV dataset, we consider that runs were unsuccessful if more than $20\%$ of the total frames could not be tracked either due to the delayed map initialization or complete tracking failures.
On the TUM monoVO dataset, we disable the keyframe culling of ORB-VO/SLAM and our systems within the start- and end-segment of each trajectory where the ground-truth data is available.
This prevents the lack of keyframes when computing the alignment error on these segments.
All images were preloaded into memory, but not rectified or photometrically calibrated beforehand.
All results were obtained in real-time on a laptop CPU (Intel Core i7-4810MQ, quad-core at 2.8 GHz with 15GB RAM).

\vspace{-0.3em}
\subsection{Results}
\label{subsec:results}
\begin{figure*}[t]
 \centering
 \vspace{5pt}
 \includegraphics[width=\textwidth]{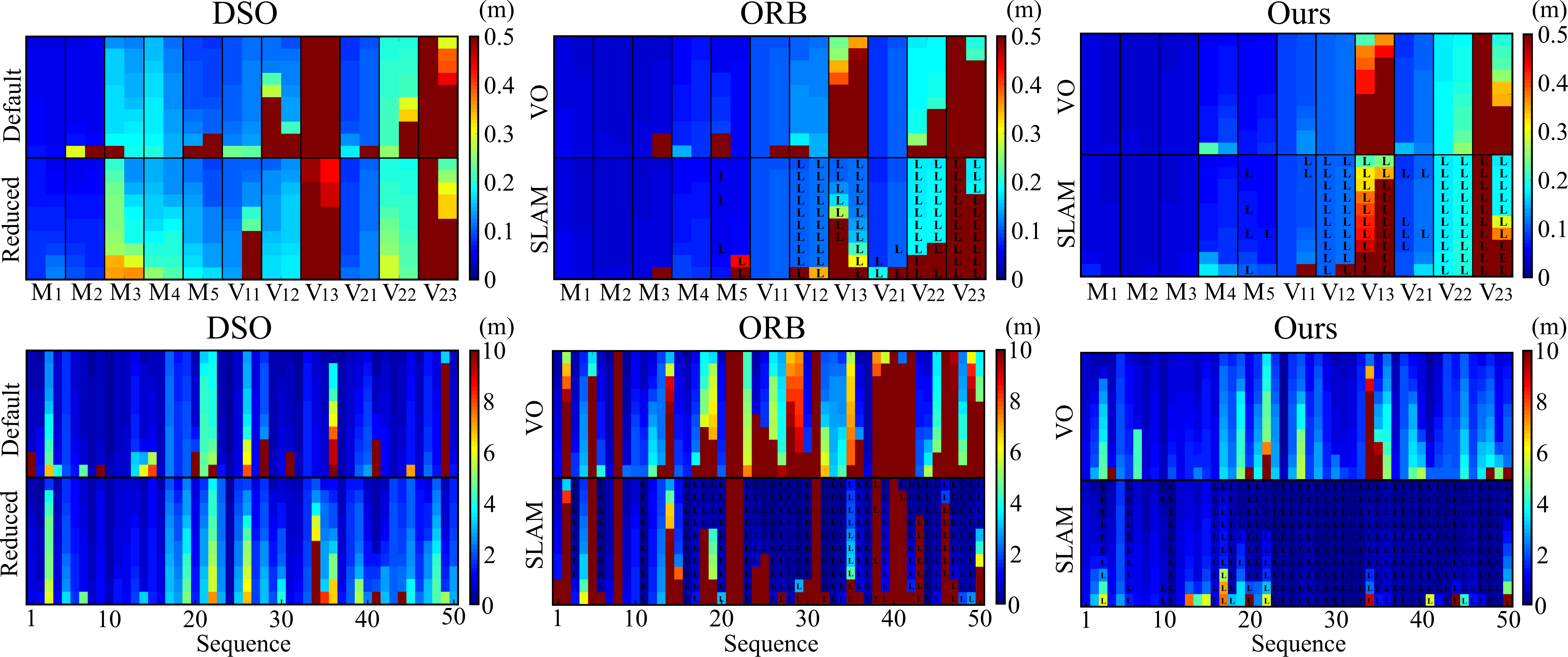}
\caption{Each of the colored blocks represent the final error value from each trial on each sequence. The letter `L' indicates the presence of one or more loop closure links. \textbf{Top row:} Absolute trajectory errors $e_\text{ate}$ [m] on the EuRoC MAV dataset. $\text{M}_{1-5}$: Machine Hall sequences. $\text{V}_{11-13, \ 21-23}$: Vicon Room 1 and 2 sequences. \textbf{Bottom row:} Alignment errors $e_\text{align}$ [m] on the TUM monoVO dataset.   }
\label{fig:error_map}
\vspace{-5pt}
\end{figure*}

Fig. \ref{fig:error_map} shows the error values for each sequence of both datasets, and their median values are reported in the supplementary material available at our public source-code repository.
The aggregated results are given in Fig. \ref{fig:cumulative} in the form of cumulative error plots indicating how many runs have yielded error values below a certain level.
On both datasets, we make three common observations:
First, DSO-reduced is more robust than DSO-default, which is most likely due to the faster tracking speed, as shown in Tab. \ref{tab:times}.
We observe similar accuracy between the two settings on the EuRoC MAV dataset, but higher accuracy with DSO-default on the TUM monoVO dataset.
Second, loop closing in ORB-SLAM improves the performance.
It increases accuracy on the TUM monoVO dataset and robustness on both datasets.
Third, due to the non-deterministic nature of real-time multi-threading, the occurrence of loop closure is not necessarily consistent on each sequence (see Fig. \ref{fig:error_map}).

On the EuRoC MAV dataset, DSO (both default and reduced) yields the lowest accuracy.
It was also reported in \cite{dso} that DSO was less accurate than ORB-VO on the same dataset.
However, we were unable to reproduce their exact results, in particular those showing that DSO was more robust in real-time.
Our systems (both VO and SLAM) and ORB-SLAM show very similar performance on this dataset,   except for \textit{V1{\_}03{\_}difficult} where ORB-SLAM outperforms our SLAM system.
This is because in ORB-SLAM, a loop closure reintroduces the detected map points into the local map and robustifies the tracking.
On the other hand, our direct tracking does not reuse the detected map points from the loop closures, so its robustness is unaffected.

The comparison between DSO-reduced and our VO system indicates that the feature-based pose refinement (Section \ref{subsec:fb_refinement}) and local mapping (Section \ref{subsec:fb_mapping_loop}) can improve the accuracy, even without loop closure.
Fig. \ref{fig:evolution} shows this effect over time on two of the sequences of the EuRoC MAV dataset. 
Notice the increased amount of drift when we do not reuse the map points that leave the FOV.
We also note that even with loop closure, the local geometric BA is still important because the pose graph optimization alone does not guarantee the optimal reconstruction of the local environment.

\begin{figure}[t]
\centering
\vspace{5pt}
 \includegraphics[width=0.48\textwidth]{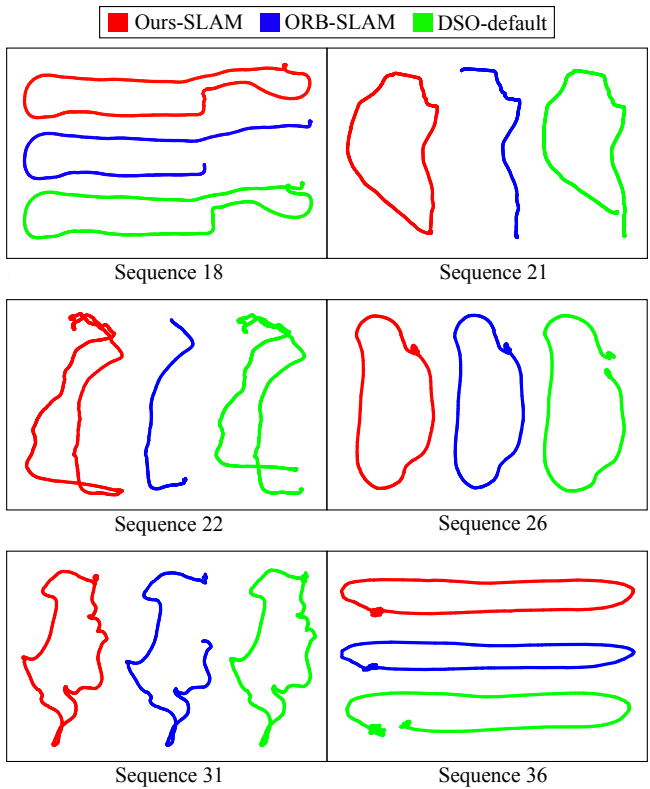}
\caption{\textbf{[TUM monoVO]} Sample trajectories estimated with the median accuracy. All sequences have the ground-truth trajectories that start and end at the same positions. For several sequences, ORB-SLAM repeatedly loses tracking and DSO suffers from consistent drifts. On the other hand, our system tracks the entire trajectories and close the loops most of the time.}
\label{fig:trajectories}
\vspace{-1em}
\end{figure}

On the TUM monoVO dataset, DSO (both default and reduced) is significantly more robust than ORB-SLAM/VO, which is similar to the results reported in \cite{dso,dso-vs-orb}.
Our VO system achieves very similar performance to DSO-reduced, as none of the final trajectories are affected by the feature-based module.
This is because more than 75\% of the total keyframes have collinear covisibility links in all sequences (see Section. \ref{subsec:does_it_always_improve}), which is in contrast to the EuRoC MAV dataset where the opposite is true.
Fig. \ref{fig:cumulative} and \ref{fig:error_map} show that the loop closure significantly reduces the alignment errors for our SLAM system in the majority of the sequences.
As a result, our SLAM system achieves the best overall performance across both datasets.
Fig. \ref{fig:trajectories} compares the estimated trajectories on some of the TUM monoVO sequences. 
Note how ORB-SLAM fails in the middle of some sequences and DSO accumulates drift, while our SLAM system completes the whole sequences and closes loops. 

{\renewcommand{\arraystretch}{1.2}%
\begin{table}[t]
  \vspace{5pt}
  \caption{\textbf{[EuRoC MAV]} Dropped frames percentage and tracking times. The two smallest values are highlighted in bold. *This is the median of the median results in each sequence.}
  \vspace{-5pt}
  \centering
  \medskip\noindent
\begin{tabularx}{0.48\textwidth}{|>{\centering\arraybackslash}p{1.6cm}||Y|Y|Y||Y|Y|Y|}
\cline{2-7}
 \multicolumn{1}{c|}{}& \multicolumn{3}{c||}{Dropped frames (\%)} & \multicolumn{3}{c|}{Tracking Times (ms)} \\
\cline{2-7}
 \multicolumn{1}{c|}{}
 & Med*
 & Mean 
 & Std 
 & Med*
 & Mean 
 & Std\\
\cline{2-7}
\hline
ORB-VO & $0.95$ & $1.56$ &$ 1.61$ & $22.09$ & $25.46$ & $8.62$ \\
\hline
ORB-SLAM & $1.54$ & $2.14$& $1.57$& $22.74$ & $26.47$ & $9.48$\\
\hline
DSO-default & $0.81$ & $1.16$ & $\mathbf{1.07}$& $7.13$ & $9.66$ &  $17.33$\\
\hline
DSO-reduced & $\mathbf{0.28}$ & $\mathbf{0.74}$& $\mathbf{1.00}$ & $\mathbf{2.60}$ & $\mathbf{4.07}$ &  $\mathbf{7.40}$\\
\hline
Ours-VO & $\mathbf{0.31}$ & $1.02$ & $1.48$ & $\mathbf{4.62}$ & $\mathbf{6.19}$ &  $\mathbf{8.62}$\\
\hline
Ours-SLAM & $0.32$ & $\mathbf{0.97}$& $1.41$& $4.69$ & $6.23$ &  $9.48$  \\
\hline
\end{tabularx}
\label{tab:times}
\end{table}
}

\begin{figure}[t]
 \centering
 \includegraphics[width=0.4\textwidth]{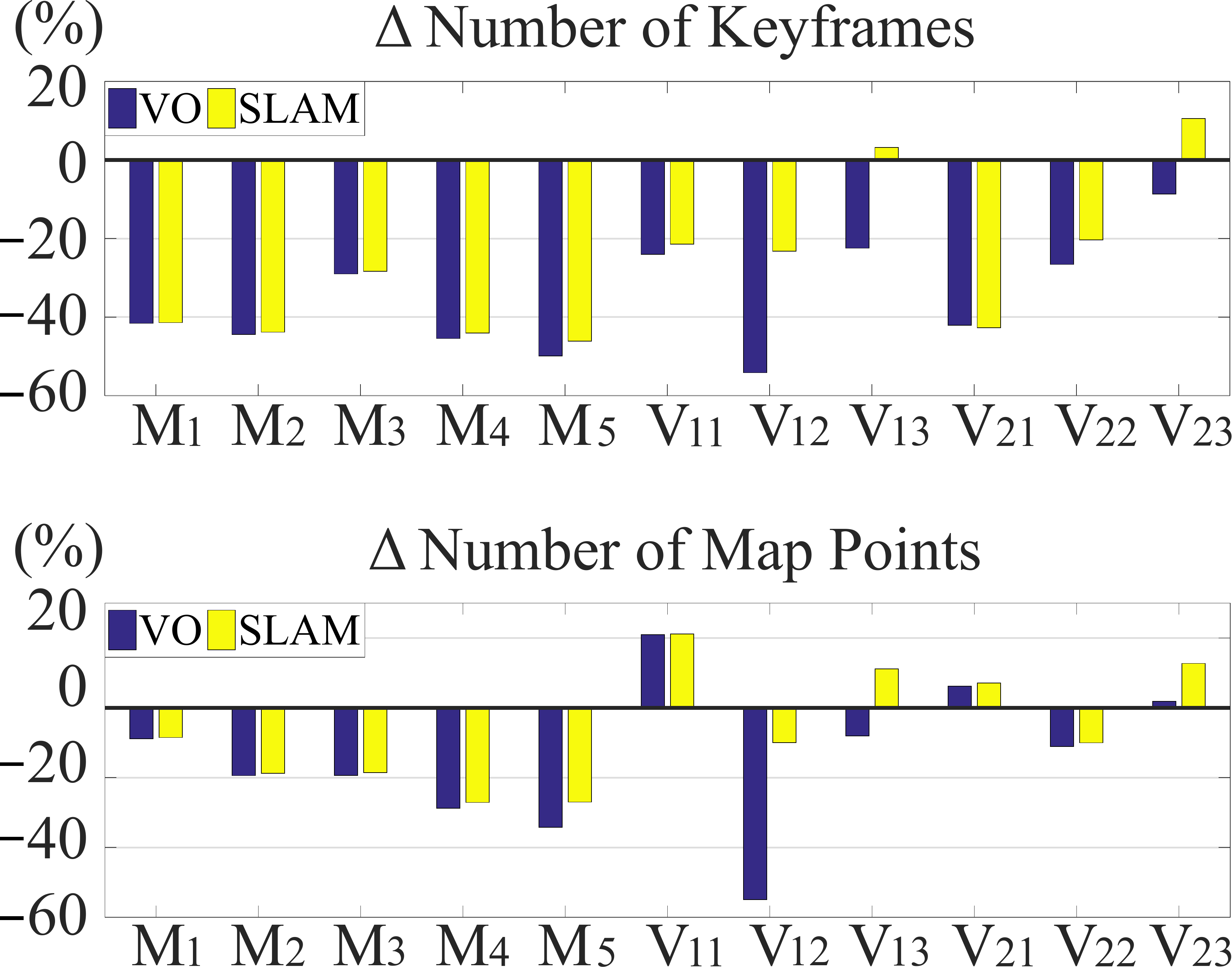}
\caption{\textbf{[EuRoC MAV]} Percentage difference in the number of total KFs and map points of our VO/SLAM compared to ORB-VO/SLAM, respectively. $\text{M}_{1-5}$: Machine Hall sequences. $\text{V}_{11-13, \ 21-23}$: Vicon Room 1 and 2 sequences.}
\label{fig:euroc_reduction}
\vspace{-10pt}
\end{figure}

Tab. \ref{tab:times} summarizes the statistics of dropped frames and tracking times\footnote{This is defined as $t_{i}-t_{i-1}$ where $t_{i}$ and $t_{i-1}$ are the two timestamps at which the tracking thread first processes frame $i$ and $i-1$, respectively.} on the EuRoC MAV dataset.
It can be seen that DSO-reduced and both our systems have lower frame drops and faster tracking speed than the rest.
This shows the advantage of direct tracking, which eliminates the need to perform feature extraction and matching for every frame. 
In Fig. \ref{fig:euroc_reduction},  we further show how much percentage of keyframes and map points are reduced in our systems compared to ORB-VO/SLAM. 
We observe average 27\% (up to 46\%) keyframes reduction and average 6\% (up to 27\%) map points reduction for the SLAM system.
This suggests that our system is relatively more scalable than ORB-SLAM.
Note that the sparsity of the feature-based keyframes and map points is what enabled the efficient map reuse on both local and global scale.
Without relying on the sparse keypoints, it would require prohibitively more computation to maintain a global covisibility graph with the large number of map points in the direct module.

\section{CONCLUSIONS}
\label{sec:conclusion}
In this paper, we proposed a loosely-coupled semi-direct method for real-time monocular SLAM.
Our system consists of two modules running in parallel.
One module uses a direct method to track new frames fast and robustly with respect to a local semi-dense map.
The other module uses the resulting points and pose estimates as prior to build a globally consistent sparse feature-based map.
We have shown on two public datasets that our method outperforms the state-of-the-art in terms of tracking accuracy and robustness.

\bibliographystyle{IEEEtran} 
\bibliography{root}

\end{document}